\documentclass[10pt]{article}

\usepackage[utf8]{inputenc}
\usepackage[T1]{fontenc}
\usepackage{times}
\usepackage[margin=1in]{geometry}
\usepackage{graphicx}
\usepackage{booktabs}
\usepackage{amsmath,amssymb}
\usepackage{hyperref}
\usepackage{xcolor}
\usepackage{caption}
\usepackage{subcaption}
\usepackage[numbers,sort]{natbib}
\usepackage{microtype}
\usepackage{placeins}  %

\definecolor{linkblue}{HTML}{1A5DAB}
\hypersetup{
  colorlinks=true,
  linkcolor=linkblue,
  citecolor=linkblue,
  urlcolor=linkblue,
}

\makeatletter
\long\def\author#1{\gdef\@author{#1}}
\makeatother

\title{GigaPath-Flash and GigaTIME-Flash:\\Efficient Pathology Foundation Models
\\for Whole-Slide and Tumor Microenvironment Analysis}

\author{
\begin{minipage}{0.98\textwidth}
\centering
Naoto Usuyama\textsuperscript{1,*}, Jeya Maria Jose Valanarasu\textsuperscript{1,*}, Sicong Yao\textsuperscript{1,*}, Hanwen Xu\textsuperscript{1,2,*},\\
Jaspreet Bagga\textsuperscript{1}, Guanghui Qin\textsuperscript{1}, Robert E. Kramer\textsuperscript{3}, Cliff Wong\textsuperscript{1}, Soohee Lee\textsuperscript{5}, Hao Qiu\textsuperscript{1}, \\
Theodore Zhengde Zhao\textsuperscript{1}, Racheli Ben Shimol\textsuperscript{4}, Angela Crabtree\textsuperscript{4}, Kevin Matlock\textsuperscript{4}, \\
Eduardo Alejandro Lozano Garcia\textsuperscript{1}, Naiteek Sangani\textsuperscript{1}, Alberto Santamaria-Pang\textsuperscript{1},\\
Maximilian Rokuss\textsuperscript{1}, Yashna Hasija\textsuperscript{1}, Naisargi Manishkumar Patel\textsuperscript{1}, Jason Entenmann\textsuperscript{1}, \\
Alexandra Q. Bartlett\textsuperscript{4}, Bill J. Wright\textsuperscript{5}, Bernard A. Fox\textsuperscript{4}, Brian Piening\textsuperscript{3,4} \\
 Sheng Zhang\textsuperscript{1}, Sheng Wang\textsuperscript{2}, Tristan Naumann\textsuperscript{1}, Carlo Bifulco\textsuperscript{3,4}, Hoifung Poon\textsuperscript{1}
\par\vspace{0.9em}
\normalsize
\textsuperscript{1}Microsoft Research \\
\textsuperscript{2}Paul G. Allen School of Computer Science and Engineering, University of Washington\\
\textsuperscript{3}Providence Genomics\\
\textsuperscript{4}Earle A. Chiles Research Institute, Providence Cancer Institute\\
\textsuperscript{5}Providence Research Network
\end{minipage}
}

\date{}

\begin{document}
\maketitle
\renewcommand{\thefootnote}{\fnsymbol{footnote}}%
\footnotetext[1]{These authors contributed equally.}%
\renewcommand{\thefootnote}{\arabic{footnote}}%

\vspace{-1.5em}
\begin{center}
\makebox[0pt][r]{%
  \href{https://aka.ms/GigaPath-Flash}{aka.ms/GigaPath-Flash}%
  \hspace{0.8em}%
}%
~~
\makebox[0pt][l]{%
  \hspace{0.8em}%
  \href{https://aka.ms/GigaTIME-Flash}{aka.ms/GigaTIME-Flash}%
}
\end{center}
\vspace{-0.5em}

\begin{abstract}
Foundation models have emerged as a driving force in computational pathology, with the potential to transform cancer diagnosis, prognosis, and treatment selection by learning transferable representations from large-scale histopathology data. Over the past few years, a flourishing landscape of pathology foundation models has emerged, spanning different data scales and sources, model architectures, and downstream applications. However, most pretrained models operate only at the image-tile level and are released under restrictive licenses, and many remain computationally expensive. Given the vast number of whole-slide images generated each year, this computational burden poses a major barrier to large-scale slide-level clinical and research applications.
Here, we introduce GigaPath-Flash and GigaTIME-Flash, efficient models designed to democratize whole-slide pathology AI and spatial proteomics prediction. GigaPath-Flash combines a 22M-parameter ViT-S tile encoder with a 21M-parameter LongNet slide encoder, both pretrained on large-scale real-world histopathology data. The compact tile encoder is distilled from the billion-parameter GigaPath (ViT-g) teacher, transferring its representational quality into a backbone an order of magnitude smaller, and this shared encoder underpins both GigaPath-Flash and GigaTIME-Flash. Despite its compact size, GigaPath-Flash retains 97\% of GigaPath's average slide-level performance while using 50× less compute. GigaTIME-Flash extends GigaPath-Flash to predict the tumor immune microenvironment directly from routine H\&E images, replacing the CNN backbone of the original GigaTIME model. It surpasses the original CNN-based GigaTIME in prediction quality while running 6× faster and using 8× less GPU memory.
Together with GigaPath and GigaTIME, these models form an open-weight, Apache-2.0-licensed family pretrained on large-scale real-world clinical data. By releasing all models and weights, we provide accessible and efficient building blocks for advancing computational pathology, immuno-oncology, and precision health.

\end{abstract}

\newpage

\section{Introduction}
\label{sec:intro}

Histopathology provides a direct view of tissue phenotype and is a central source of information across cancer biology, translational research, drug development, and patient care. A whole-slide image (WSI) captures information across multiple spatial scales, ranging from cellular morphology and local tissue composition to slide-wide tissue architecture and the organization of the tumor microenvironment. These patterns are relevant to a broad range of research and clinical applications, including tumor classification, biomarker discovery, patient stratification, prognosis, treatment-response prediction, and spatial characterization of immune states.

With the rise of foundation models as a powerful paradigm for learning transferable representations from large-scale data~\citep{usuyama2025biomedical}, computational pathology has increasingly embraced large-scale pretraining to capture the rich and diverse information encoded in WSIs. Over the past few years, a growing number of pathology foundation models have demonstrated strong performance across cancer types and downstream tasks, establishing large-scale pretraining as an important foundation for computational pathology.

\begin{figure}[!b]
  \centering
  \includegraphics[width=\textwidth]{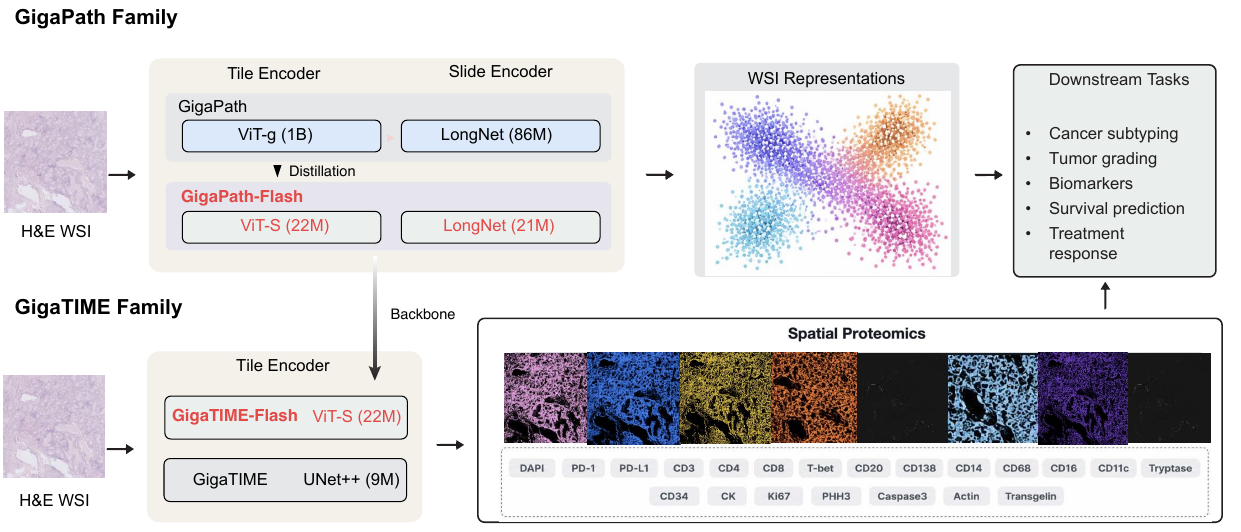}
  \caption{
    Overview of the GigaPath/GigaTIME model family.
    GigaPath-Flash provides efficient tile and slide encoders at 22M parameters.
    GigaTIME predicts spatial proteomics from H\&E, scaling from CNN (9M) to ViT backbones.
  }
  \label{fig:family_overview}
\end{figure}

Despite this progress, four barriers limit their broader use. First, many models are computationally expensive, a challenge amplified in whole-slide analysis because the tile encoder must be applied thousands of times per slide. Second, most are pretrained only at the tile level and therefore do not directly learn slide-wide tissue architecture or long-range spatial context. Third, many are trained primarily on public, research-curated cohorts rather than large-scale real-world diagnostic data. Finally, restrictive licenses can limit their use and adaptation in both academic and commercial research. Together, these considerations motivate efficient, slide-aware models pretrained on real-world data and released under permissive licenses.

We introduce GigaPath-Flash and GigaTIME-Flash, efficient models that address these four barriers and extend the GigaPath/GigaTIME family~\citep{xu2024gigapath,valanarasu2026multimodal} (Figure~\ref{fig:family_overview}). GigaPath-Flash combines a 22M-parameter ViT-S tile encoder with a 21M-parameter LongNet slide encoder that together embed a whole slide using $\sim$49.5$\times$ fewer FLOPs than the original GigaPath, both pretrained on large-scale real-world histopathology data. The tile encoder is distilled from the original GigaPath ViT-g (1B) teacher, and GigaTIME-Flash reuses this distilled encoder to predict spatial protein expression and characterize the tumor immune microenvironment directly from routine H\&E images.

These models retain or improve upon the performance of original model, while being efficient in both GPU memory and inference time. GigaPath-Flash outperforms models with up to 31$\times$ more parameters on slide-level classification benchmarks while requiring 37$\times$ less compute. GigaTIME-Flash surpasses the original CNN-based GigaTIME while running approximately 6$\times$ faster and using 8$\times$ less GPU memory. Together with GigaPath and GigaTIME, these models form an open-weight, Apache~2.0-licensed family spanning tile-level representation learning, whole-slide analysis, and spatial proteomics prediction.

\begin{table}[!h]
\centering
\caption{GigaPath and GigaTIME model family.}
\label{tab:family}
\small
\begin{tabular}{lllll}
\toprule
Model & Type & Architecture & License & Release \\
\midrule
GigaPath & Whole-Slide FM & ViT-g (1B) + LongNet (86M) & Apache-2.0 & Nature, 2024 \\
GigaPath-Flash & Whole-Slide FM & ViT-S (22M) + LongNet (21M) & Apache-2.0 & \it{New} \\
\midrule
GigaTIME & Spatial Proteomics & CNN U-Net++ (9M) & Apache-2.0 & Cell, 2026 \\
GigaTIME-Flash & Spatial Proteomics & ViT-S (22M) + Decoder (2M) & Apache-2.0 & \it{New} \\
\bottomrule
\end{tabular}
\end{table}

\section{GigaPath-Flash}
\label{sec:gigapath}

\subsection{Model}

GigaPath-Flash is an efficient foundation model that turns a gigapixel pathology slide into a spatially contextualized whole-slide representation.
It has two components: a ViT-S/16 tile encoder (22M parameters) that maps each $224\times224$\,px tile to a 384-dimensional embedding, and a 12-layer LongNet~\citep{ding2023longnet} slide encoder (21M parameters, 384 dimensions) that jointly contextualizes all tile embeddings via dilated attention, scaling linearly with the number of tiles.
The tile encoder is distilled from the frozen GigaPath ViT-g (1B) teacher with a DINOv2~\citep{oquab2023dinov2} objective on whole-slide images from the Providence real-world cohort, transferring the representational capacity of the billion-parameter model into a backbone an order of magnitude smaller. We follow the standard DINOv2 recipe but omit the KoLeo regularization term, which we found to destabilize training when distilling into this compact student.
The slide encoder is then pretrained on the resulting tile features with a masked autoencoding objective, learning to predict a tile's representation from the rest of the slide and thereby encoding each location in the context of the whole.

\subsection{Slide-Level Benchmarks}

We evaluate pathology foundation models on two public slide-level benchmarks.
PANDA comprises H\&E-stained prostate biopsy slides labeled with ISUP grade
groups~\citep{bulten2022panda}; we formulate this as six-class ordinal
classification (grades 0--5) and report quadratic-weighted Cohen's kappa (QWK).
For EBRAINS~\citep{roetzer2022ebrains}, we evaluate 30-class fine-grained brain
tumor subtyping and report balanced accuracy to account for class imbalance.

\begin{figure}[!htb]
  \centering
  \includegraphics[width=\columnwidth]{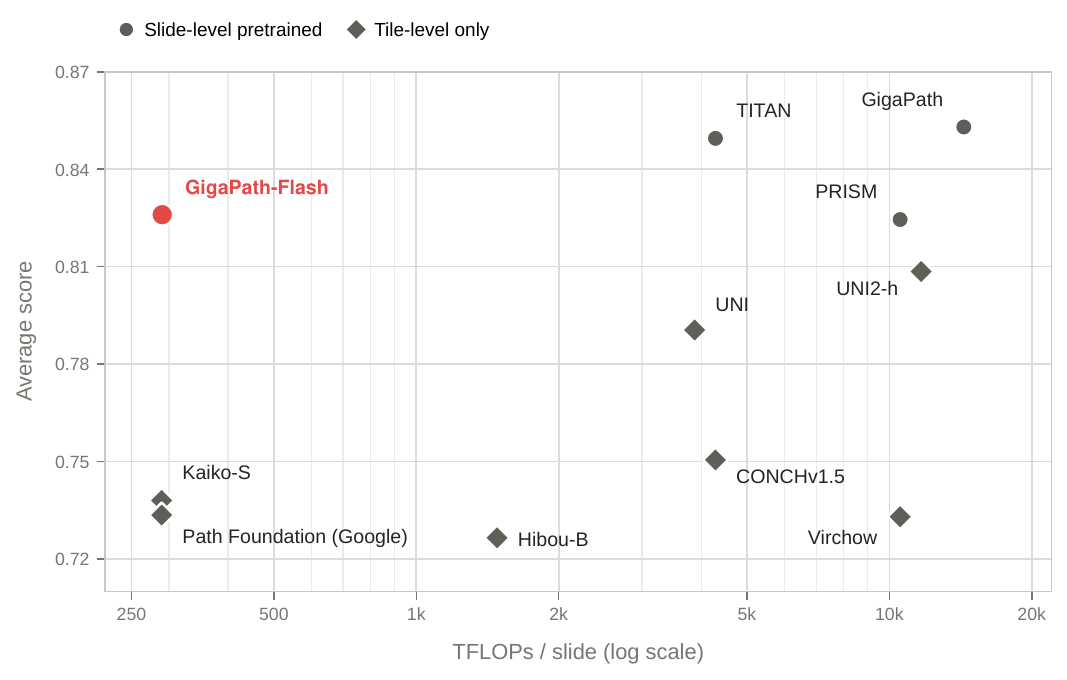}
  \caption{
    Efficiency--performance trade-off on whole-slide benchmarks.
    TFLOPs/slide ($x$-axis, log scale) vs.\ average score ($y$-axis).
    Circles indicate slide-level pretrained models; diamonds indicate tile-level-only models aggregated with ABMIL.
  }
  \label{fig:pareto}
\end{figure}

Whole-slide images are tiled at 20$\times$ magnification into non-overlapping
$256\times256$\,px tiles using our tiling pipeline \cite{xu2024gigapath}. For TITAN~\citep{ding2025titan},
adjacent tiles are stitched to form the $512\times512$\,px inputs expected by
the model. Slide-level foundation models use their native slide
encoders, whereas tile-level models are aggregated with attention-based
multiple-instance learning (ABMIL)~\citep{ilse2018attention}. We estimate TFLOPs per slide using the largest EBRAINS slide, which yields 31{,}469 tiles at $256\times256$\,px and 7{,}964 tiles at $512\times512$\,px under this tiling setup. Per-slide FLOPs are computed as the per-tile encoder FLOPs multiplied by the number of tiles, plus the FLOPs of the native slide encoder or ABMIL aggregator. FLOPs are measured using PyTorch's \texttt{FlopCounterMode}. For FLOP measurement only, the LongNet-based slide encoders in GigaPath and GigaPath-Flash use an equivalent unfused attention implementation. For the TensorFlow-based Path Foundation model, tile-encoder FLOPs are derived from weight-tensor shapes and verified against an architecture-matched PyTorch implementation, with exact agreement.

We created a fixed train/validation/test split for each dataset and used the
same split across all models. Downstream models are trained for five epochs,
and the final-epoch checkpoint is evaluated without validation-based model
selection. We report a single run for each dataset. Because these custom splits
differ from official or previously published protocols, results should be
interpreted as controlled comparisons within our protocol.

\begin{table}[!htb]
  \centering
  \caption{Slide-level benchmark results and model characteristics. TFLOPs/slide are computed for an EBRAINS slide containing 31{,}469 $256\times256$-px tiles; relative TFLOPs use GigaPath-Flash as $1.0\times$. Average score is the unweighted mean of PANDA QWK and EBRAINS balanced accuracy.}
  \label{tab:slide_benchmarks}
  \resizebox{\textwidth}{!}{%
  \begin{tabular}{lclrrccc}
    \toprule
    Model & Whole-Slide FM & License & \shortstack{TFLOPs/\\slide} & \shortstack{Relative\\TFLOPs} & \shortstack{PANDA\\QWK} & \shortstack{EBRAINS\\Bal. Acc.} & \shortstack{Average\\score} \\
    \midrule
    GigaPath-Flash & \checkmark & Apache-2.0 & 290.3 & $1.0\times$ & 0.947 & 0.705 & 0.8260 \\
    GigaPath~\citep{xu2024gigapath} & \checkmark & Apache-2.0 & 14{,}367.3 & $49.5\times$ & 0.965 & 0.741 & 0.8530 \\
    PRISM~\citep{shaikovski2024prism} & \checkmark & CC BY-NC-ND 4.0 & 10{,}536.6 & $36.3\times$ & 0.945 & 0.704 & 0.8245 \\
    TITAN~\citep{ding2025titan} & \checkmark & CC BY-NC-ND 4.0 & 4{,}289.8 & $14.8\times$ & 0.942 & 0.757 & 0.8495 \\
    UNI~\citep{chen2024uni} & --- & CC BY-NC-ND 4.0 & 3{,}874.1 & $13.3\times$ & 0.935 & 0.646 & 0.7905 \\
    UNI2-h~\citep{mahmoodlab2025uni2} & --- & CC BY-NC-ND 4.0 & 11{,}673.2 & $40.2\times$ & 0.939 & 0.678 & 0.8085 \\
    Hibou-B~\citep{nechaev2024hibou} & --- & Apache-2.0 & 1{,}481.6 & $5.1\times$ & 0.928 & 0.525 & 0.7265 \\
    Kaiko-S\citep{kaiko2024kaiko} & --- & Kaiko NC & 289.5 & $1.0\times$ & 0.934 & 0.542 & 0.7380 \\
    Path Foundation (Google)~\citep{lai2023pathfoundation} & --- & HAI-DEF ToU & 289.5 & $1.0\times$ & 0.932 & 0.535 & 0.7335 \\
    \bottomrule
  \end{tabular}
  }

  \vspace{3pt}
  \begin{minipage}{\textwidth}
  \footnotesize Kaiko NC: Kaiko Non-Commercial Public License. HAI-DEF ToU: Health AI Developer Foundations Terms of Use.\\.
  \end{minipage}
\end{table}

Across PANDA and EBRAINS, slide-level pretrained models formed the highest-performing group in the comparison (Table~\ref{tab:slide_benchmarks} and Figure~\ref{fig:pareto}). GigaPath-Flash achieved the lowest inference cost among these whole-slide models while outperforming every tile-level-only baseline on both tasks, placing it favorably in the efficiency--performance trade-off. The license comparison further shows that GigaPath-Flash provides an Apache-2.0-licensed option within this high-performing group of slide-level pretrained models. These results position GigaPath-Flash as an compact and permissively available whole-slide foundation model with strong performance across distinct slide-level classification tasks.

\FloatBarrier
\section{GigaTIME-Flash}
\label{sec:gigatime}

\subsection{Model}

GigaTIME follows a UNet++ architecture \citep{zhou2018unet++} which takes in a H\&E patch as an input and predicts 21 channel multiplex immunofluorescence maps across protein markers -  as output. UNet++ has a convolutional encoder and decoder framework with skip connections. In the supplementary material of \cite{valanarasu2026multimodal}, we had shown that when we replace the encoder of GigaTIME with GigaPath weights and train it for mIF prediction we do get an improvement in performance with a trade-off in the number of parameters and inference time which would be inefficient while doing any population-scale inference with the model.

As GigaPath-Flash is an efficient distilled model compared to the full GigaPath encoder, we introduce GigaTIME-Flash, where the encoder is initialized from the distilled GigaPath-Flash ViT-small checkpoint and paired with a lightweight convolutional decoder for H\&E-to-mIF translation. The encoder is a 12-layer ViT-small backbone with patch size $16$, embedding dimension $384$, and $6$ attention heads. For a $256 \times 256$ H\&E input tile, this produces a $16 \times 16$ grid of patch tokens. Skip features are extracted from the 4th, 6th, 9th, and 12th transformer blocks. The deepest feature from block 12 initializes the decoder, while features from blocks 9, 6, and 4 are injected into successive decoder stages through learned transposed-convolution skip projections.
The decoder follows a U-Net-style convolutional design. The block-12 token map is reshaped into a spatial feature map and passed through four decoder stages with channel dimensions $384 \rightarrow 192 \rightarrow 96 \rightarrow 48 \rightarrow 24$. Each decoder stage contains two $3 \times 3$ convolution layers with ReLU activations, followed by bilinear upsampling by a factor of 2. The skip paths project encoder features using transposed convolutions: the block-9 feature is projected to 192 channels, the block-6 feature to 96 channels, and the block-4 feature to 48 channels before being added to the corresponding decoder activations. A final $1 \times 1$ convolution maps the 24-channel decoder output to the 21 mIF output channels.

To adapt the encoder while limiting catastrophic forgetting, we fine-tune using LoRA~\citep{hu2021lora} adapters on the transformer attention \texttt{qkv} and output projection \texttt{proj} modules. The LoRA configuration uses rank $r=8$, scaling parameter $\alpha=16$, and dropout $0.1$. Non-LoRA encoder parameters are frozen, while the LoRA adapters and the convolutional decoder are trainable. Overall, GigaTIME-Flash contains $23{,}806{,}559$ parameters. The model is trained for 21-channel mIF prediction using BCEDice loss, Adam optimization with learning rate $10^{-4}$ and weight decay $10^{-4}$, and a cosine annealing learning-rate schedule with minimum learning rate $10^{-5}$. GigaTIME-Flash was trained for 300 epochs on a NVIDIA A100 GPU node with batch size of 64 per GPU on the same GigaTIME training data for fairness \citep{valanarasu2026multimodal}.

\begin{figure}[!htb]
  \centering
  \includegraphics[width=\columnwidth]{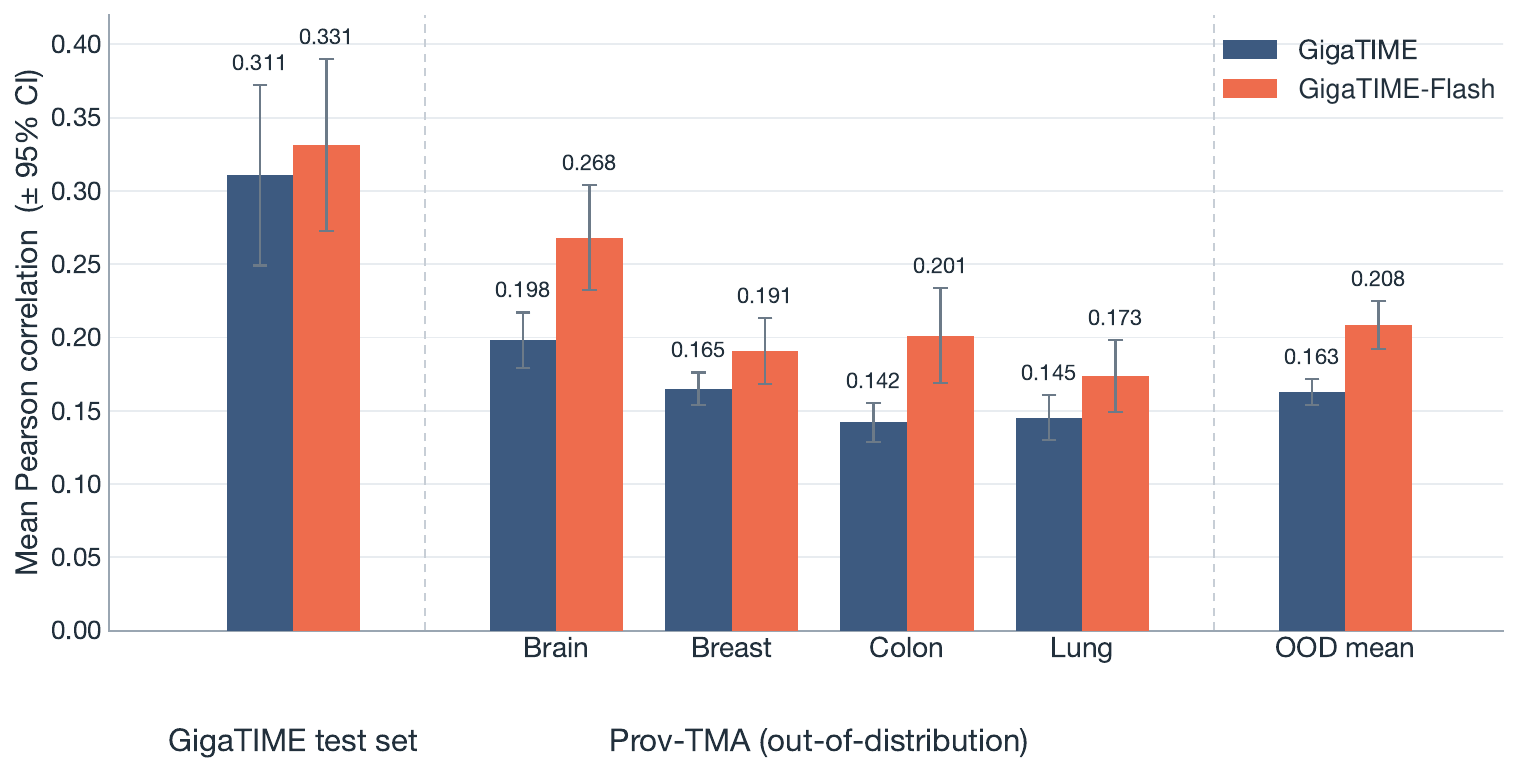}
  \caption{
Mean windowed pearson correlation for GigaTIME and GigaTIME-Flash on the GigaTIME test set and four out-of-distribution Prov-TMA cohorts. Error bars indicate 95\% confidence intervals.
  }
  \label{fig:gigatime_bars}
\end{figure}

\subsection{Results}

We test the performance of GigaTIME-Flash on 2 datasets - 1) the original GigaTIME test set consisting of high quality registered 9,204 tiles of size $512\times512$ from 5 tissues from 5 patients patients of lung adenocarcinoma (LUAD) cohort. 2) Providence tissue micro-array (Prov-TMA) dataset which is collected across 4 different cancer types - Brain, Breast, Colon, and Lung (LUSC). For each organ site, tissue microarrays were constructed from whole-slide sections and include samples from approximately 10 to 20 patients per organ site each with a tissue micro-array, providing an out-of-distribution evaluation across diverse tumor types.

Following the original GigaTIME evaluation protocol, we quantify performance using the Pearson correlation coefficient computed over non-overlapping $8\times8$ pixel windows. For both the predicted protein activation map and the corresponding ground-truth mIF image, the activation values within each $8\times8$ window are aggregated to form a coarse activation matrix. The Pearson correlation is then computed between the predicted and ground-truth activation matrices for each marker, and the average correlation is reported. An $8\times8$ window corresponds approximately to the spatial extent of individual cells, making the metric more biologically meaningful while reducing sensitivity to minor registration errors and pixel-level noise. Additional details regarding this evaluation protocol are provided in~\cite{valanarasu2026multimodal}.

\begin{figure}[!htb]
  \centering
  \includegraphics[width=\columnwidth]{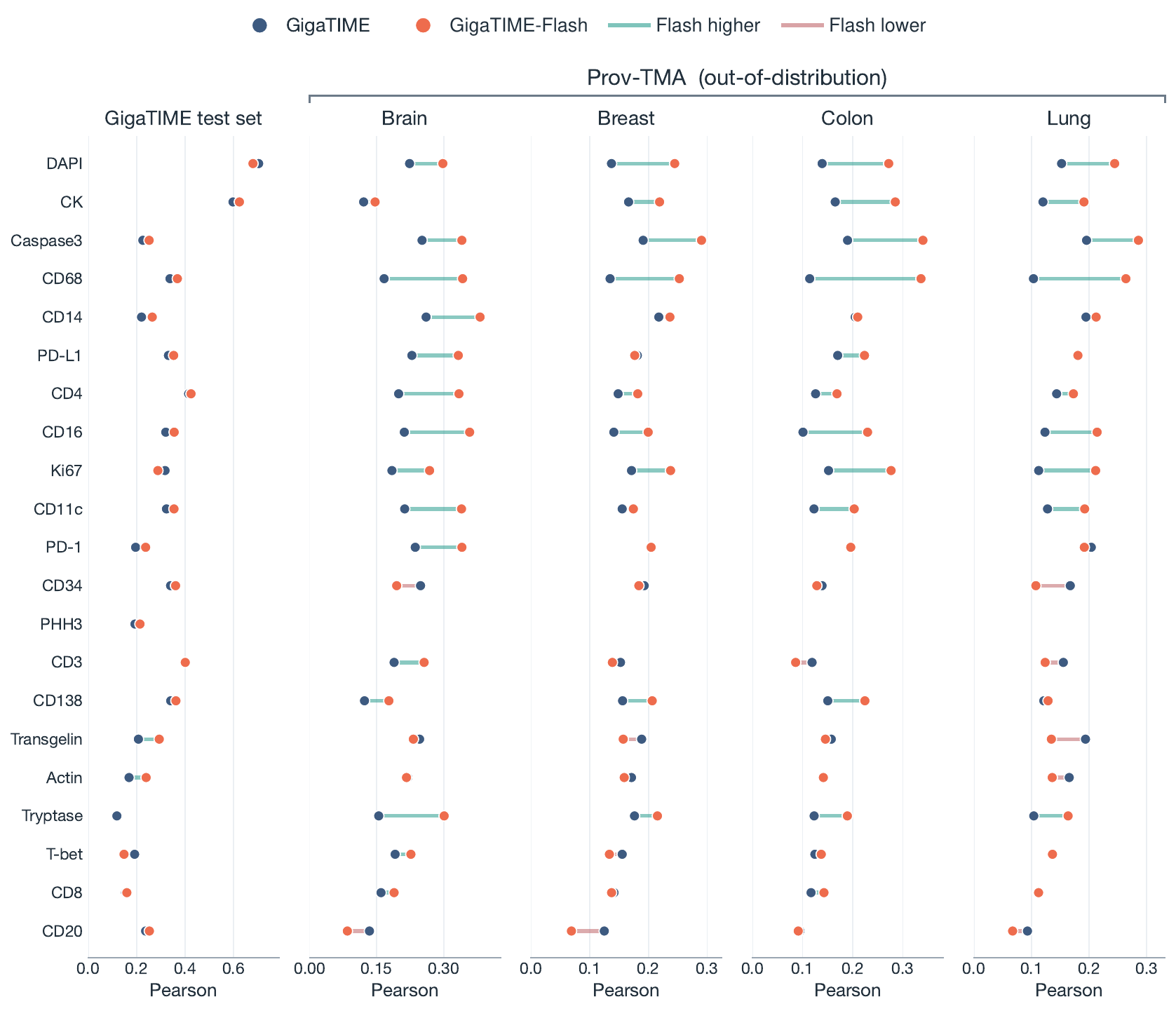}
  \caption{
   Protein marker level spatial proteomics prediction performance across in-distribution GigaTIME test set and out-of-distribution Prov-TMA cohorts.
    The panel displays all 21 output channels; PHH3 channel is unavailable for OOD dataset as it was not collected. Points indicate the correlation for each model and channel; connecting segments show paired differences, with teal indicating higher performance for GigaTIME-Flash and pink indicating lower performance.
  }
  \label{fig:gigatime_markers}
\end{figure}

GigaTIME-Flash achieves a higher mean Pearson correlation than the original GigaTIME across the in-distribution test set and all four out-of-distribution (OOD) cohorts (Figure~\ref{fig:gigatime_bars}) as it is trained with a foundation model as the encoder following the trend we reported in supplementary material of \cite{valanarasu2026multimodal}.
The difference is modest on the GigaTIME test set but more pronounced across the OOD cohorts, suggesting that the foundation-model backbone improves transfer to previously unseen cohorts.

The marker-level comparison shows that these gains are broad but not uniform (Figure~\ref{fig:gigatime_markers}).
GigaTIME-Flash improves the prediction of nuclear and epithelial patterns, several myeloid markers, and markers of proliferation and apoptosis.
In contrast, the original CNN remains competitive for some vascular, stromal, and sparse lymphoid patterns, including CD34, Transgelin, Actin, and CD20. Overall, the results support GigaTIME-Flash as a more generalizable model for broad tumor-microenvironment profiling.

\subsection{Efficiency}

\begin{figure}[!htb]
  \centering
  \includegraphics[width=\columnwidth]{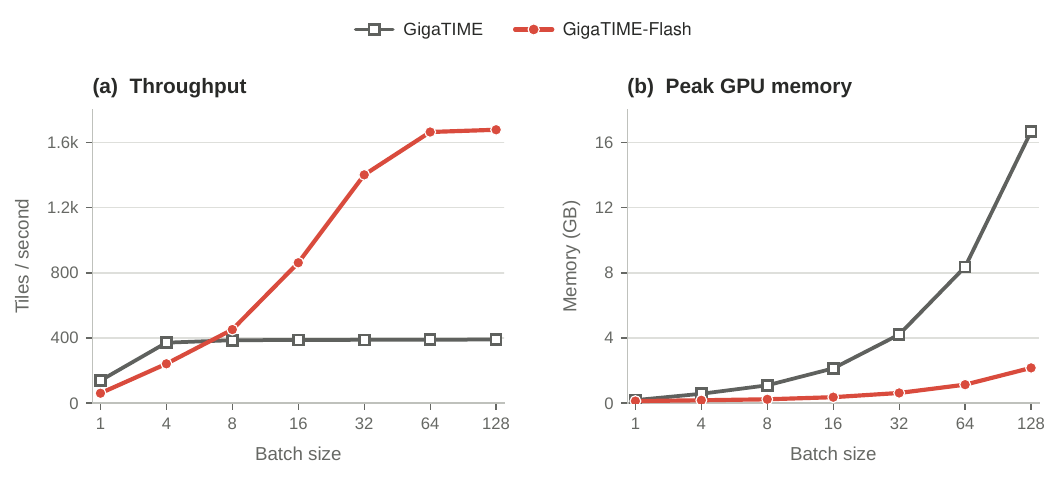}
  \caption{
    GigaTIME efficiency scaling.
    Left: throughput (tiles/sec) vs.\ batch size.
    Right: peak GPU memory (GB) vs.\ batch size.
    GigaTIME-Flash throughput increases from 60.3 to 1{,}679.2 tiles/sec, while GigaTIME plateaus at approximately 390 tiles/sec.
    At batch size 128, GigaTIME-Flash uses 2.16\,GB of peak GPU memory compared with 16.68\,GB for GigaTIME.
    All measurements were obtained on an NVIDIA A100 GPU.
  }
  \label{fig:gigatime_efficiency}
\end{figure}

GigaTIME-Flash achieves substantially higher inference efficiency than the original CNN-based GigaTIME despite having a larger number of parameters (Figure~\ref{fig:gigatime_efficiency}). For a $256\times256$ image tile, GigaTIME-Flash requires only 14.9 GFLOPs compared to 69.1 GFLOPs for GigaTIME, corresponding to a $\sim$4.6$\times$ reduction in computational cost. This improvement stems from architectural differences: GigaTIME performs dense convolutional operations with nested skip connections over high-resolution feature maps throughout the network, whereas GigaTIME-Flash carries out most computation on its latent token representation and relies on a lightweight convolutional decoder for reconstruction.

Furthermore, the ViT-based backbone exhibits significantly better hardware utilization on modern GPUs. As batch size increases, large matrix multiplication operations become increasingly efficient and can be parallelized effectively across GPU cores, allowing throughput to continue scaling. In contrast, the convolution-heavy UNet++ architecture quickly saturates available GPU resources, resulting in minimal throughput gains beyond moderate batch sizes.

As shown in Figure~\ref{fig:gigatime_efficiency}, GigaTIME-Flash scales from 60.3 to 1,679.2 tiles/s as the batch size increases from 1 to 128, whereas GigaTIME plateaus at approximately 390 tiles/s. At the same time, GigaTIME-Flash requires only 2.16~GB of peak GPU memory at batch size 128 compared to 16.68~GB for GigaTIME, representing nearly an 8$\times$ reduction in memory footprint.

These efficiency gains are particularly important for population-scale virtual biomarker generation. A single whole-slide image typically contains thousands of tissue tiles, and clinical cohorts often comprise hundreds of thousands of slides. The combination of higher throughput and substantially lower memory consumption enables significantly faster deployment across large real-world datasets, reduces GPU infrastructure requirements, and lowers the overall computational cost of generating virtual multiplex protein maps at population scale.

\FloatBarrier
\section{Limitations}
\label{sec:limitations}

We note several limitations. Our slide-level evaluation covers only two public classification benchmarks and does not assess transfer to other tasks such as survival analysis, retrieval, or treatment response predictions. It also uses one custom split, a single run per model, and a common five-epoch training recipe, which does not capture variability across splits or random seeds and may not optimally tune every model; scores are therefore not directly comparable with those reported under other protocols. GigaPath-Flash trades some absolute performance for efficiency, and efficiency measurements on a single NVIDIA A100 may not generalize across hardware or implementations. GigaTIME-Flash was evaluated with one LoRA configuration and a fixed 21-marker panel in cohorts with limited numbers of independent patients. Its in-silico predictions and windowed Pearson correlation do not establish cell-level accuracy or clinical utility. Broader multi-institutional and prospective validation across tasks, cohorts, scanners, and patient subgroups is needed before deployment.

\section{Conclusion}
\label{sec:conclusion}

We introduced GigaPath-Flash and GigaTIME-Flash, additions to the GigaPath/GigaTIME model family that provide an efficient foundation for whole-slide representation learning and spatial tumor immune microenvironment prediction. GigaPath-Flash retains competitive performance on the evaluated slide-level benchmarks with substantially reduced compute, while GigaTIME-Flash improves prediction quality over the original CNN-based GigaTIME with lower inference time and GPU memory requirements. Although broader evaluation and external validation remain necessary, these results highlight the potential of efficient foundation models to balance predictive performance with practical efficiency. By lowering computational barriers, we hope these models will enable research across increasingly large and diverse cohorts and help accelerate the shift from isolated tile-level analysis toward context-aware whole-slide modeling and scalable spatial characterization of the tumor immune microenvironment.

{\small
\bibliographystyle{unsrtnat}
\bibliography{references}

\begin{thebibliography}{17}
\providecommand{\natexlab}[1]{#1}
\providecommand{\url}[1]{\texttt{#1}}
\expandafter\ifx\csname urlstyle\endcsname\relax
  \providecommand{\doi}[1]{doi: #1}\else
  \providecommand{\doi}{doi: \begingroup \urlstyle{rm}\Url}\fi

\bibitem[Usuyama et~al.(2025)Usuyama, Wong, Zhang, Naumann, and Poon]{usuyama2025biomedical}
Naoto Usuyama, Cliff Wong, Sheng Zhang, Tristan Naumann, and Hoifung Poon.
\newblock Biomedical natural language processing in the era of large language models.
\newblock \emph{Annual Review of Biomedical Data Science}, 8\penalty0 (1):\penalty0 471--490, 2025.
\newblock \doi{10.1146/annurev-biodatasci-103123-095406}.

\bibitem[Xu et~al.(2024)Xu, Usuyama, Bagga, Zhang, Rao, Naumann, Wong, Gero, Gonz{\'a}lez, Gu, Xu, Wei, Wang, Ma, Wei, Yang, Li, Gao, Rosemon, Bower, Lee, Weerasinghe, Wright, Robicsek, Piening, Bifulco, Wang, and Poon]{xu2024gigapath}
Hanwen Xu, Naoto Usuyama, Jaspreet Bagga, Sheng Zhang, Rajesh Rao, Tristan Naumann, Cliff Wong, Zelalem Gero, Javier Gonz{\'a}lez, Yu~Gu, Yanbo Xu, Mu~Wei, Wenhui Wang, Shuming Ma, Furu Wei, Jianwei Yang, Chunyuan Li, Jianfeng Gao, Jaylen Rosemon, Tucker Bower, Soohee Lee, Roshanthi Weerasinghe, Bill~J. Wright, Ari Robicsek, Brian Piening, Carlo Bifulco, Sheng Wang, and Hoifung Poon.
\newblock A whole-slide foundation model for digital pathology from real-world data.
\newblock \emph{Nature}, 630\penalty0 (8015):\penalty0 181--188, 2024.

\bibitem[Valanarasu et~al.(2026)Valanarasu, Xu, Usuyama, Kim, Wong, Argaw, Shimol, Crabtree, Matlock, Bartlett, et~al.]{valanarasu2026multimodal}
Jeya Maria~Jose Valanarasu, Hanwen Xu, Naoto Usuyama, Chanwoo Kim, Cliff Wong, Peniel Argaw, Racheli~Ben Shimol, Angela Crabtree, Kevin Matlock, Alexandra~Q Bartlett, et~al.
\newblock Multimodal ai generates virtual population for tumor microenvironment modeling.
\newblock \emph{Cell}, 189\penalty0 (2):\penalty0 386--400, 2026.

\bibitem[Ding et~al.(2023)Ding, Ma, Dong, Zhang, Huang, Wang, Zheng, and Wei]{ding2023longnet}
Jiayu Ding, Shuming Ma, Li~Dong, Xingxing Zhang, Shaohan Huang, Wenhui Wang, Nanning Zheng, and Furu Wei.
\newblock Longnet: Scaling transformers to 1,000,000,000 tokens.
\newblock \emph{arXiv preprint arXiv:2307.02486}, 2023.

\bibitem[Oquab et~al.(2023)Oquab, Darcet, Moutakanni, Vo, Szafraniec, Khalidov, Fernandez, Haziza, Massa, El-Nouby, et~al.]{oquab2023dinov2}
Maxime Oquab, Timoth{\'e}e Darcet, Th{\'e}o Moutakanni, Huy Vo, Marc Szafraniec, Vasil Khalidov, Pierre Fernandez, Daniel Haziza, Francisco Massa, Alaaeldin El-Nouby, et~al.
\newblock {DINOv2}: Learning robust visual features without supervision.
\newblock \emph{arXiv preprint arXiv:2304.07193}, 2023.
\newblock \doi{10.48550/arXiv.2304.07193}.

\bibitem[Bulten et~al.(2022)Bulten, Kartasalo, Chen, Str{\"o}m, Pinckaers, Nagpal, Cai, Steiner, van Boven, Vink, et~al.]{bulten2022panda}
Wouter Bulten, Kimmo Kartasalo, Po-Hsuan~Cameron Chen, Peter Str{\"o}m, Hans Pinckaers, Kunal Nagpal, Yuannan Cai, David~F Steiner, Hester van Boven, Robert Vink, et~al.
\newblock Artificial intelligence for diagnosis and gleason grading of prostate cancer: the panda challenge.
\newblock \emph{Nature Medicine}, 28:\penalty0 154--163, 2022.

\bibitem[Roetzer-Pejrimovsky et~al.(2022)Roetzer-Pejrimovsky, Moser, Atli, Vogel, Mercea, Prihoda, Gelpi, Haberler, H{\"o}ftberger, Hainfellner, et~al.]{roetzer2022ebrains}
Thomas Roetzer-Pejrimovsky, Anna-Christina Moser, Baran Atli, Clemens~Christian Vogel, Petra~A. Mercea, Romana Prihoda, Ellen Gelpi, Christine Haberler, Romana H{\"o}ftberger, Johannes~A. Hainfellner, et~al.
\newblock The digital brain tumour atlas, an open histopathology resource.
\newblock \emph{Scientific Data}, 9\penalty0 (1):\penalty0 55, 2022.

\bibitem[Ding et~al.(2025)Ding, Wagner, Song, Chen, Lu, Zhang, Vaidya, Jaume, Shaban, Kim, et~al.]{ding2025titan}
Tong Ding, Sophia~J. Wagner, Andrew~H. Song, Richard~J. Chen, Ming~Y. Lu, Andrew Zhang, Anurag~J. Vaidya, Guillaume Jaume, Muhammad Shaban, Ahrong Kim, et~al.
\newblock A multimodal whole-slide foundation model for pathology.
\newblock \emph{Nature Medicine}, 31:\penalty0 3749--3761, 2025.

\bibitem[Ilse et~al.(2018)Ilse, Tomczak, and Welling]{ilse2018attention}
Maximilian Ilse, Jakub Tomczak, and Max Welling.
\newblock Attention-based deep multiple instance learning.
\newblock In \emph{Proceedings of the International Conference on Machine Learning}, pages 2127--2136, 2018.

\bibitem[Shaikovski et~al.(2024)Shaikovski, Casson, Severson, Zimmermann, Wang, Kunz, Retamero, Oakley, Klimstra, Kanan, et~al.]{shaikovski2024prism}
George Shaikovski, Adam Casson, Kristen Severson, Eric Zimmermann, Yi~Kan Wang, Jeremy~D. Kunz, Juan~A. Retamero, Gerard Oakley, David Klimstra, Christopher Kanan, et~al.
\newblock {PRISM}: A multi-modal generative foundation model for slide-level histopathology.
\newblock \emph{arXiv preprint arXiv:2405.10254}, 2024.

\bibitem[Chen et~al.(2024)Chen, Ding, Lu, Williamson, Jaume, Song, Chen, Zhang, Shao, Shaban, et~al.]{chen2024uni}
Richard~J Chen, Tong Ding, Ming~Y Lu, Drew~FK Williamson, Guillaume Jaume, Andrew~H Song, Bowen Chen, Andrew Zhang, Daniel Shao, Muhammad Shaban, et~al.
\newblock Towards a general-purpose foundation model for computational pathology.
\newblock \emph{Nature Medicine}, 2024.

\bibitem[{Mahmood Lab}(2025)]{mahmoodlab2025uni2}
{Mahmood Lab}.
\newblock {UNI2-h}.
\newblock \url{https://huggingface.co/MahmoodLab/UNI2-h}, 2025.
\newblock Model release.

\bibitem[Nechaev et~al.(2024)Nechaev, Pchelnikov, and Ivanova]{nechaev2024hibou}
Dmitry Nechaev, Alexey Pchelnikov, and Ekaterina Ivanova.
\newblock Hibou: A family of foundational vision transformers for pathology.
\newblock \emph{arXiv preprint arXiv:2406.05074}, 2024.

\bibitem[{kaiko.ai} et~al.(2024){kaiko.ai}, Aben, de~Jong, Gatopoulos, K{\"a}nzig, Karasikov, Lagr{\'e}, Moser, van Doorn, and Tang]{kaiko2024kaiko}
{kaiko.ai}, Nanne Aben, Edwin~D. de~Jong, Ioannis Gatopoulos, Nicolas K{\"a}nzig, Mikhail Karasikov, Axel Lagr{\'e}, Roman Moser, Joost van Doorn, and Fei Tang.
\newblock Towards large-scale training of pathology foundation models.
\newblock \emph{arXiv preprint arXiv:2404.15217}, 2024.

\bibitem[Lai et~al.(2023)Lai, Ahmed, Vijay, Jaroensri, Loo, Vyawahare, Agarwal, Jamil, Matias, Corrado, et~al.]{lai2023pathfoundation}
Jeremy Lai, Faruk Ahmed, Supriya Vijay, Tiam Jaroensri, Jessica Loo, Saurabh Vyawahare, Saloni Agarwal, Fayaz Jamil, Yossi Matias, Greg~S. Corrado, et~al.
\newblock Domain-specific optimization and diverse evaluation of self-supervised models for histopathology.
\newblock \emph{arXiv preprint arXiv:2310.13259}, 2023.
\newblock \doi{10.48550/arXiv.2310.13259}.

\bibitem[Zhou et~al.(2018)Zhou, Rahman~Siddiquee, Tajbakhsh, and Liang]{zhou2018unet++}
Zongwei Zhou, Md~Mahfuzur Rahman~Siddiquee, Nima Tajbakhsh, and Jianming Liang.
\newblock Unet++: A nested u-net architecture for medical image segmentation.
\newblock In \emph{International workshop on deep learning in medical image analysis}, pages 3--11. Springer, 2018.

\bibitem[Hu et~al.(2022)Hu, Shen, Wallis, Allen-Zhu, Li, Wang, Wang, and Chen]{hu2021lora}
Edward~J. Hu, Yelong Shen, Phillip Wallis, Zeyuan Allen-Zhu, Yuanzhi Li, Shean Wang, Lu~Wang, and Weizhu Chen.
\newblock {LoRA}: Low-rank adaptation of large language models.
\newblock In \emph{Proceedings of the International Conference on Learning Representations}, 2022.

\end{thebibliography}
}

\end{document}